\documentclass[10pt,a4paper,oneside]{article}
\pdfoutput=1
\usepackage[english]{babel}
\usepackage{placeins}
\usepackage[utf8]{inputenc}
\usepackage{color,multiobjective}
\usepackage{amsmath,amssymb}
\usepackage{algorithmic}
\usepackage{algorithm}
\usepackage{balance}
\usepackage{ctable}
\usepackage{fancyhdr}
\usepackage{marvosym}
\usepackage{a4wide, subfigure}
\usepackage[small, bf ,hang, nooneline]{caption}

\newcommand{\otoprule}{\midrule[\heavyrulewidth]}

\newtheorem{definition}{Definition}

\def\Real{{\mathbb{R}}}

\def\GDp{\mbox{$\mbox{\sf GD}_p$}{}}

\def\IGDp{\mbox{$\mbox{\sf IGD}_p$}{}}

\def\Mf#1{\mbox{$\mbox{\small\sffamily ND}_f({ #1 }, \preceq)$}}

\def\Mf#1{\mbox{$\mbox{\small\sffamily ND}_f({ #1 }, \preceq)$}}

\begin{document}

\setcounter{page}1

\thispagestyle{empty}
\begin{center}
\vspace*{1cm}

{\Large {\bf Averaged Hausdorff Approximations of Pareto Fronts\vspace*{0.3cm}

  based on Multiobjective Estimation \vspace*{0.3cm}
  
  of Distribution Algorithms}} 
\vspace*{1.5cm}

{\Large 2015}

\vspace*{1.5cm}

\begin{tabular}{cc}
Luis Mart{\'\i} \hspace*{1cm}& Christian Grimme\\
Departamento de Engenharia El\'{e}trica \hspace*{1cm}& Information Systems and Statistics\\
PUC Rio de Janeiro, Brazil \hspace*{1cm}& University of M{\"u}nster, Germany\\
lmarti@ele.puc-rio.br \hspace*{1cm}& christian.grimme@wi.uni-muenster.de\\[3.5ex]

Pascal Kerschke\hspace*{1cm}&Heike Trautmann\\
Information Systems and Statistics\hspace*{1cm}&Information Systems and Statistics\\
University of M{\"u}nster, Germany\hspace*{1cm}&University of M{\"u}nster, Germany\\
kerschke@uni-muenster.de\hspace*{1cm}&trautmann@wi.uni-muenster.de\\[5.5ex]
\end{tabular}

G{\"u}nter Rudolph\\
Department of Computer Science\\
TU Dortmund University, Germany\\
guenter.rudolph@tu-dortmund.de

\end{center}
\vspace*{1.5cm}

\paragraph{Abstract\\}
\noindent In the a posteriori approach of multiobjective optimization the Pareto front is approximated by a finite set of solutions in the objective space. The quality of the approximation can be measured by different indicators that take into account the approximation's closeness to the Pareto front and its distribution along the Pareto front. In particular, the averaged Hausdorff indicator prefers an almost uniform distribution.
An observed drawback of multiobjective estimation of distribution algorithms (MEDAs) is that - as common for randomized metaheuristics - the final population usually is not uniformly distributed along the Pareto front. Therefore, we propose a postprocessing strategy which consists of applying the averaged Hausdorff indicator to the complete archive of generated solutions after optimization in order to select a uniformly distributed subset of nondominated solutions from the archive.
In this paper, we put forward a strategy for extracting the above described subset. The effectiveness of the proposal is contrasted in a series of experiments that involve different MEDAs and filtering techniques.

\newpage

\section{Introduction}

Many real-world optimization problems involve more than one goal to be optimized and are known as multiobjective optimization problems (MOPs), i.e. a
set of objective functions $f_1(\vec{x}),\ldots,$ $f_M(\vec{x})$ should be jointly optimized; formally,
\begin{equation}\label{eqn:mop-problem}
\min\ \vec{F}(\vec{x})= \left(\,f_1(\vec{x}), \ldots, f_M(\vec{x})\,\right);\ \vec{x}\in\set{S}\,;
\end{equation}
where $\set{S}\subseteq\mathbb{R}^n$ is known as the feasible set and could be expressed as a set of
restrictions over the decision or search space $\mathbb{R}^n$.
The image set $\set{O}\subseteq\mathbb{R}^M$ of $\set{S}$ produced by the vector-valued function
$\vec{F}(\cdot)$  is called feasible objective set or criterion set.
The solution to this type of problem is a set of trade-off
points. The optimality of a solution can be expressed in terms of the Pareto dominance relation.
\begin{definition}[Pareto dominance]\label{def:domination}
In the optimization problem (\ref{eqn:mop-problem}) and having
$\vec{x},\vec{y}\in\set{S}$, $\vec{x}$ is said to dominate $\vec{y}$ (expressed as $\vec{x}\dom\vec{y}$)
if $\forall j=1,\ldots M$:
$f_j(\vec{x})\leq f_j(\vec{y})$ and $\exists i\in\{1,\ldots, M\}$: $f_i(\vec{x})< f_i(\vec{y})$.
The \emph{non-dominated subset} $\set{A}^\ast$ of set $\set{A}\subseteq\set{S}$ is
defined as
\begin{equation}
\set{A}^\ast=\left\{ \vec{x}\in\set{A}
\left|\not{\exists}\vec{x}'\in\set{A}:\vec{x}'\dom\vec{x}\right.\right\}. \nonumber
\end{equation}
\end{definition}
The solution of (\ref{eqn:mop-problem}) is $\set{S}^\ast$, the non-dominated subset of $\set{S}$.
$\set{S}^\ast$ is known as the \emph{efficient set} or \emph{Pareto-optimal set}
\cite{dag-2008:mo-inter-evo}. Its image in objective space is known as the \emph{Pareto-optimal
front}, $\set{O}^\ast$.
As finding the explicit formulation of $\set{S}^{\ast}$ is often impossible, generally, an
algorithm solving (\ref{eqn:mop-problem}) yields a discrete non-dominated set, $\set{P}^\ast$,
that approximates $\set{S}^{\ast}$. The image of $\set{P}^\ast$ in objective set, $\set{PF}^\ast$,
is known as the \emph{non-dominated front}.

A broad range of heuristic and metaheuristic approaches has been used to address MOPs
\cite{dag-2008:mo-inter-evo}. Of these, evolutionary multiobjective  optimization algorithms (EMOAs)
\cite{coello-2007:eas-solving-mops} have been found to be a competent approach in a wide variety of
application domains. Alternatively, multiobjective estimation of distribution algorithms (MEDAs) were introduced which aim at learning the problem structure and characteristics along the run. In this paper, we will experimentally  investigate how representative MEDAs perform compared to classical EMOAs and how especially MEDAs can be improved after the run by means of a postprocessing approach in order to get more equally spaced solutions on the final Pareto front approximation.

The crucial task is how to measure the performance in the multiobjective setting, i.e. how to asses the relation of $\set{PF}^\ast$ to $\set{O}^\ast$.
Several performance indicators have been proposed including the hypervolume indicator or  the R indicators, see \cite{Zitzler02,ZTL03} for an overview. Each indicator concentrates on special desired characteristics of the front approximation while one frequently discussed aim is that elements  of $\set{PF}^\ast$ should be evenly spread along the true Pareto front in order to present an unbiased solution set to the decision maker (e.g. \cite{Deb02}). On the application side, specifically optimization systems or online control could profit from such approximations such that changes from one solution to the other are quite moderate and of equal extent.
In \cite{emmerich:13} an approach for evolving Pareto front approximations with uniform gap is presented. However, most indicators do not address this  issue.

The $\Delta_p$ indicator introduced in \cite{SELC12} specifically favors approximations with the desired characteristic based on averaged Hausdorff distances between $\set{PF}^\ast$ and $\set{O}^\ast$.
Several algorithmic approaches have been introduced so far (e.g. \cite{GRST11, RTSS13, DRST13} to generate quite equally spaced Pareto front approximations but all of them are focusing on integrating the $\Delta_p$ based subset selection into the EMOA. In this paper, we present a strategy to be applied posterior to an approximation run to extract a most uniformly distributed subset from the archive of all nondominated solutions generated within the course of the used EMOA.

The remainder of this paper is organized as follows: In Section \ref{sec:bg}
the methodological background is provided regarding multiobjective estimation of distribution algorithms and the postprocessing strategy.
Experimental results are  presented  and discussed in Section \ref{sec:algoExp}. Conclusions are drawn in Section \ref{sec:conc} supplemented by an outlook on further research directions.

\section{Methodology}\label{sec:bg}

\subsection{Multiobjective Estimation of Distribution Algorithms}

The inclusion of learning as part of the search process can improve the performance of evolutionary multiobjective optimization algorithms \cite{corne-2008:talk}. It can be argued that learning would allow to grasp the characteristics of the problem being solved and, hence, explore the search space in a more efficient manner. There are some approaches that perform this task by providing hybrid evolutionary/machine learning methods, like, for example, the learnable evolution model (LEM) \cite{michalski-2000:lem}. Other approaches attempt to infer the fitness landscape of the problem in order to find promising search directions \cite{garrett-2008:mo-fit-land,verel-2010:mofitland,humeau-2013:paradiseo-mo}.
Another alternative for carrying out this task is to resort to what has been denominated as estimation of distribution algorithms (EDAs) \cite{lozano-2005:edas}. This is because of EDAs capacity of learning the problem structure. EDAs replace the application of evolutionary operators in the offspring generation process with the creation of a statistical model of the fittest elements of the population in a process known as model-building. This model is then sampled to produce new elements. Nevertheless, the so called multiobjective EDAs (MEDAs) \cite{pelikan-2006:medas} have not lived up to their a priori expectations. This can be attributed to the fact that most MEDAs have limited themselves to transforming single-objective EDAs into a multiobjective formulation by including an existing multiobjective fitness assignment function.

Bosman and Thierens \cite{bosman2005MIDEA} proposed the successful multiobjective mixture-based iterated density estimation algorithms (MIDEAs). They also proposed a novel Pareto-based and diversity-preserving fitness assignment function. MIDEA considered several types of probabilistic models for both discrete and continuous problems. A mixture of univariate distributions and a mixture of tree distributions were used for discrete variables. A mixture of univariate Gaussian models and a mixture of multivariate Gaussian factorizations were applied for continuous variables. Different variants of MIDEAs implement this in a particular form. For example, naive MIDEA the naive Bayes algorithm.

It can also be argued that the covariance matrix adaptation evolution strategies (CMA-ES) \cite{hansen-2003:cma-es} are also EDAs, as they construct an abstract model of the population and then sample it to produce new individuals. CMA-ES provides a method for updating the
covariance matrix of the multivariate normal mutation distribution used in an evolution strategy
\cite{beyer-2002:evol-strats}. The covariance matrix describes the pairwise dependencies between the variables in the
distribution. Adaptation of the covariance matrix is equivalent to learning a second-order model of the
underlying objective function. The multi-objective CMA-ES (MO-CMA-ES) \cite{igel2007MOCMAES} is the extrapolation to the multi-objective domain.

It has been pointed out that the current model-building algorithms of MEDAs have a set of drawbacks that would prevent those algorithms from yielding substantially better results. In particular, the tendency of MEDAs loosing population diversity has repeatedly been reported \cite{ahn-2007:diversity-preservation,yuan-2005:diversity-eda}. This situation, although already described in single-objective EDAs \cite{shapiro-2006:diversity-eda}, is particularly dramatic in the multiobjective case, as diversity and homogeneity are among the desired features of the final non-dominated set. An analysis of the results yielded by current MEDAs and their scalability with regard to the number of objectives leads to hypotheses regarding the issues that might be hampering the obtention of substantially better results with regard to other evolutionary approaches \cite{marti-2011:phd-thesis}.

The incorrect handling of data outliers is a paradigmatic example of insufficient comprehension of the nature of the model building problem. In `standard' machine learning practice, outliers are treated as noisy, inconsistent or irrelevant data. Therefore, this type of data is expected to have little influence on the model or just to be disregarded. However, that behavior is not adequate for model building. In this case, it is known beforehand that all elements in the data set should be taken into account as they represent newly discovered or candidate regions of the search space and therefore must be explored. Therefore, these instances should be at least equally represented by the model and perhaps even reinforced. This situation is further aggravated by the mating selection scheme employed in most MEDAs. The continuous selection of the best part of the population might lead to a premature homogenization of the population and, therefore, to the stagnation of the search process. In the second case, the loss of diversity can be traced back to the above-described outliers issue of model-building algorithms and also to the incorrect estimation or sampling of the model. This fact leads us back to the statement referring that model building has not been correctly acknowledged as a different problem with particular requirements.

There have been some works that propose to improve this situation by introducing model building algorithms better suited for the task. That is the case of the MIDEA algorithm family, with the introduction of the adaptive variance scaling (AVS) and the standard deviation ratio (SDR) \cite{Bosman2007}. The AVS and SDR combination helps fight the early reduction of the mixture densities variances and therefore the premature convergence and diversity loss. Another important milestone has been the introduction of the anticipated mean shift (AMS) that takes into
account the previous values of the means of the distribution to ``push'' solutions towards the Pareto-optimal front. AMS has been conjointly used with AVS in the multiobjective adapted maximum-likelihood Gaussian mixture model (MAMaLGaM-X) \cite{Bosman2010}. Similarly, the multiobjective optimization neural EDA \; (MONEDA)~\cite{marti2008MONEDA} embeds a custom-made model building algorithm \cite{marti-2011:mb-gng-orl} that is able to maintain diversity by correctly handling the outlier elements. This approach has been improved by the introduction of the match-based learning pa\-ra\-digm of adaptive resonance theory (ART) \cite{grossberg-1982:match-based} leading to the multiobjective ART EDA (MARTEDA) \cite{marti2011MARTEDA}.

Other approaches propose modifications to currently existing methods. For example, in \cite{yuan-2005:diversity-eda} a method that avoids too sudden drops (to zero) of the variances of a multivariate Gaussian model is proposed. Similarly, in \cite{brake-2007:sampling-edas} a permutation sampling is introduced that eliminates the sampling errors of UMDA \cite{Muhlenbein_and_Paas:1996}. Other approaches \cite{pena-2004:ga-eda,zhang-2005:div-loss} have tried strategies to re-inject ``fresh'' individuals that are kept on a evolutionary algorithm that is run in parallel. In spite of these efforts, the issue of population diversity and lack of homogeneity is still an open topic that must be addressed.

\subsection{Postprocessing of MEDA results}

Suppose we are particularly interested in a
finite size approximation of $\set{O}^\ast=\vec{F}(\set{S}^\ast)$
that exhibits a sufficiently good spread
and a small distance to $\set{O}^\ast$.
Previous work in the context of EMOA (see e.g.\ \cite{RTSS13})
has shown that the \emph{averaged Hausdorff distance} \cite{SELC12}
can be used to achieve this goal.

\begin{definition} Let $A, B \subset \Real^M$ be non-empty finite sets.
The value $$\Delta_p(A, B)=\max(\GDp(A,B), \IGDp(A,B)) \mbox{ with}$$
$$\GDp(A,B)=\left(\frac{1}{|A|} \sum_{a\in A} d(a,B)^p\right)^{1/p} \mbox{ and}$$
$$\IGDp(A,B)= \left(\frac{1}{|B|} \sum_{b\in B} d(b,A)^p\right)^{1/p}$$
for $p>0$
is termed the \emph{averaged Hausdorff distance} between sets $A$ and $B$ where
$\mbox{d}(u,A) := \inf\{\|u-v\|: v\in A\}$
for $u,v\in\Real^M$ and some vector norm $\|\cdot\|$.
\end{definition}

Suppose that some MEDA has generated an approximation of the Pareto front for some MOP.
Typically, this approximation does not yield a finite point set in objective space that is evenly
distributed. Therefore, we propose the following \emph{postprocessing} approach:
\begin{enumerate}
\item Run your favorite MEDA that is equipped with a tiny add-on:
store a copy of each offspring that is generated and evaluated in a file.
\item After termination of your favorite MEDA:\\
construct an evenly spaced reference front from a given approximation
of the Pareto front (e.g.\ from last population of favorite MEDA);
then feed each stored offspring into the $\Delta_p$ archive updater (see alg.\ \ref{algo:dpUpdate}) sequentially;
the final content of the archive $A$ is the desired approximation.
\end{enumerate}

For bi-objective problems the reference front $R$ can be constructed as follows:
calculate a linear interpolation from a given approximation of the Pareto front.
Since the length $L$ of the resulting polygonal line can be derived easily,
division by the number of desired archive points $m$ yields the size of the equal spacing $\delta = L/m$ which is used
to place $m$ points along the polygonal line with distance $\delta$, where the extreme points of
the discretization are moved half the length inwards the polygonal line.
After the reference front $R$ has been constructed it is used in the $\Delta_p$ archive updater to decide which point should be added to
or deleted from the archive.

An update operation can be realized as sketched in alg.\ref{algo:dpUpdate}. This naive approach takes
$\Theta(|A|\cdot ( |A|\cdot|R|\cdot M))$ time units, but there exists a quick update version
\cite{RTSS13} that only needs
$\Theta(|A|\cdot |R|\cdot M)$ time units.

\begin{algorithm}[h]
\caption{Naive $\Delta_1$-update \cite{RTSS13}\label{algo:dpUpdate} }
\begin{algorithmic}[1]
\REQUIRE archive set $A$, reference set $R$, new element $\vec{x}$
\STATE $A = \Mf{A\cup\{\vec{x}\}}$
\IF{$|A| > N_R := |R|$}
\FORALL{$a\in A$}
\STATE $h(a)=\Delta_1(A\setminus\{a\}, R)$
\ENDFOR
\STATE $A^* = \{a^* \in A: a^*=\mbox{\sf argmin}\{ h(a) : a\in A\}\}$
 \IF{$|A^*| > 1$}
   \STATE $a^{*} = \mbox{\sf argmin}\{\GDp(A\setminus\{a\}, R) : a\in A^*\}$ 
   \ENDIF
\STATE $A = A\setminus \{a^*\}$
\ENDIF
\end{algorithmic}
\end{algorithm}

The most obvious order of feeding the stored pairs ($\vec{x}$, $\vec{F(x)}$) into the archive updater is
the order of their generation. We call this the `forward update.' In this manner, many individuals
will pass the initial dominance check, so that subsequent $\Delta_p$ calculations are necessary.
Some time saving may be achieved by feeding the stored pairs into the archive update in inverted order.
We call this the `backward update.' Since points that have been generated in later iterations
of the MEDA are more likely to dominate previous points, most points from the rear of the inverted
sequence will probably not pass the initial dominance check, so that subsequent $\Delta_p$ calculations can
be avoided. Since the order of the points presented to the archive clearly affects the final outcome
of the archive, we shall compare both approaches experimentally in the next section.

\section{Experiments}\label{sec:algoExp}

\subsection{Experimental Setup}\label{subsec:ExSetup}

In order to experimentally evaluate the $\Delta_p$ archive approximation for multiple multiobjective EDAs, we implemented two offline strategies in the JAVA$\texttrademark$ jMetal framework \cite{durillo2011jmetal}: a forward $\Delta_p$-approach ($+fDP$), star\-ting with the initial population and iteratively updating with previously traced generations, whereas the backward-strategy ($+bDP$) starts with the final population. Within the $\Delta_p$-update mechanisms, a linear interpolation and a PSA-based strategy, both using the aggregated solutions of the whole run, were used for deriving a set of evenly spaced reference points. Based on that set, a solution was eventually added to the archive.

We then used well known MOPs for benchmarking \cite{GRST11}: an instance of the sphere problem with $X\!\subseteq\!\mathbb{R}^2$ and a con\-nected convex Pareto front, the DENT problem with $X\!\subseteq\!\mathbb{R}^2$ and a connected convex-concave-convex front, ZDT3 with $X\!\subseteq\!\mathbb{R}^{20}$ and a disconnected convex/concave front, as well as WFG1 with $X\!\subseteq\!\mathbb{R}^{2+4}$, also with a convex-concave front.

Then, reference fronts covered by $1,\!000$ uniformly spaced points were created in order to compare our $\Delta_p$ approximation quality for all algorithms. With the exception of WFG1, we were able to find a parametric form for exactly calculating the optimal fronts' length L (by rectification), for all benchmark problems. Given that form $1,\!000$ points were placed uniformly on the PF using distance $\Delta L = L / 1,\!000$ between points and starting with the extreme points moved inwards by $\Delta\,L\,/\,2$. In case of WFG1 \cite{huband2006review}, the neighborhood of the known Pareto set was explored using a very fine grid (per dimension of the decision space), in order to find a PF. Then $1,\!000$ well-distributed solutions among the non-dominated solutions were kept. As the number of reference solutions is by far larger than the approximated set's size, a not perfectly uniform distribution of the Pareto-optimal solutions is acceptable.

Our evaluation is based on four state-of-the-art general purpose EMOAs and four MEDAs. From the former group, we employed the NSGA2 \cite{Deb02} and SMS-EMOA \cite{beume2007sms} with standard parameter settings -- SBX crossover with $p_x\,=\,0.9$ and polynomial mutation with $p_{mut}\,=\,1/n$ -- as well as PS\-EMOA as special purpose EMOA with inherent internal clustering for generating evenly distributed solutions. For the same reason we also used a mo\-di\-fied version of NSGA2 \cite{kukkonen2006improved} -- the \textit{sequential crowding distance NSGA2} (SCD-NSGA2) -- which successively alternates between removing the individuals with the smallest crowding distance and recomputing the crowding distance values, until only $\mu$ solutions are left.

Considering the MEDAs we deal with two well-known approaches: the naive MIDEA \cite{bosman2005MIDEA} and the multiobjective CMA-ES (MO-CMA-ES) \cite{igel2007MOCMAES}. We also include MONEDA \cite{marti2008MONEDA} and MARTEDA \cite{marti2011MARTEDA} as they are supposed to have a better handling of diversity. The parameters of the MEDAs are summarized in Table \ref{tbl:algorithm-parameters}.

During our experiments, all test problems were optimized $20$ times by all algorithms, given a budget of $50,\!000$ function evaluations each and population sizes $\mu \in \{10,~20, 100\}$. The data, which were generated from the optimizers' runs, have been submitted to the offline $\Delta_p$ archivers. Both update strategies were parameterized with $p \in \{1, 2\}$.

\begin{table}[t]
\centering \scriptsize
\begin{tabular}{@{}r@{\,\ }c@{}}
\toprule
\multicolumn{2}{c}{\textbf{MARTEDA}}\\
\midrule
F2 vigilance threshold ($\rho$)                       & $0.05$\\
Initial standard deviations ($\vec{\varphi}$)         & $\vec{0.01}$\\
Selection percentile ($\alpha$)                       & $0.3$\\
$\hat{P}_t$ to $N^\ast$ ratio ($\gamma$)              & $0.5$\\
Substitution percentile ($\omega$)                    & $0.25$\\
\otoprule
\multicolumn{2}{c}{\textbf{MONEDA}}\\
\midrule
Number of initial GNG nodes ($N_0$)                   & $2$\\
Maximum edge age ($\nu_\mathrm{max}$)                 & $40$ \\
Best node learning rate ($\epsilon_\mathrm{b})$       & $0.1$\\
Neighboring nodes learning rate ($\epsilon_\mathrm{v}$)  & $0.05$\\
Insertion error decrement rate ($\delta_\mathrm{I}$)  & $0.1$\\
General error decrement rate ($\delta_\mathrm{G}$)    & $0.1$\\
Accumulated error threshold ($\rho$)                  & $0.2$\\
Selection percentile ($\alpha$)                       & $0.3$\\
$\hat{\set{P}}_t$ to $N_\mathrm{max}$ ratio ($\gamma$)      & $0.5$\\
Substitution percentile ($\omega$)                    & $0.25$\\
\otoprule
\multicolumn{2}{c}{\textbf{Naive MIDEA}}\\
\midrule
Selection percentile ($\tau$)                         & $0.3$ \\
Diversity percentile ($\delta$)                       & $15$  \\
Number of parents of a variable ($\kappa$)            & $2$   \\
Maximum number of clusters                            & $\lceil 0.5\lfloor \tau n_\mathrm{pop}\rfloor\rceil$\\
Threshold for the leader algorithm                    & $0.1$ \\
\otoprule
\multicolumn{2}{c}{\textbf{MO-CMA-ES}}\\
\midrule
Initial step size ($\sigma$)                         & $1$ \\
Damping for step-size ($d$)                           & $1 + n_\mathrm{pop} / 2$ \\
Target success rate                                   & $0.181$   \\
Cumulation time horizon                              & $2 / (n_\mathrm{pop} + 2)$\\
Covariance learning rate                             & $2 / (n_\mathrm{pop}^2 + 6)$\\
Threshold success rate                                & $0.44$ \\
\bottomrule
\end{tabular}
\caption{Parameters of the algorithms used in the experiments following the notation of the corresponding author(s).}
\label{tbl:algorithm-parameters}
\end{table}

\subsection{Results}

\begin{figure}[t]
		\centering
    \includegraphics[width=\columnwidth]{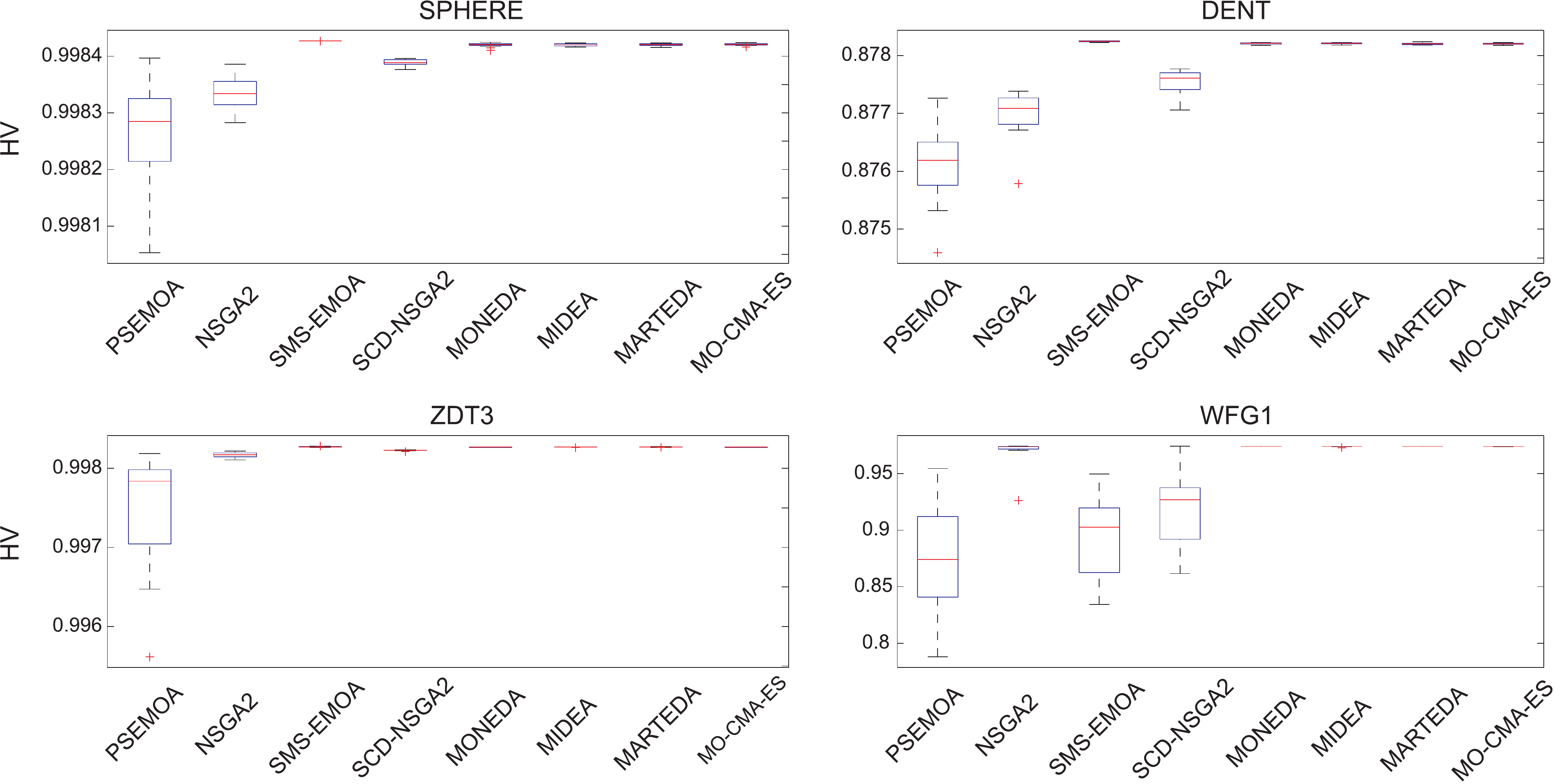}
    \caption{Comparison of MEDAs with EMOAs regarding Hypervolume.}
    \label{fig:algos_hv}
\end{figure}

\begin{figure}[t]
		\centering
    \includegraphics[width=\columnwidth]{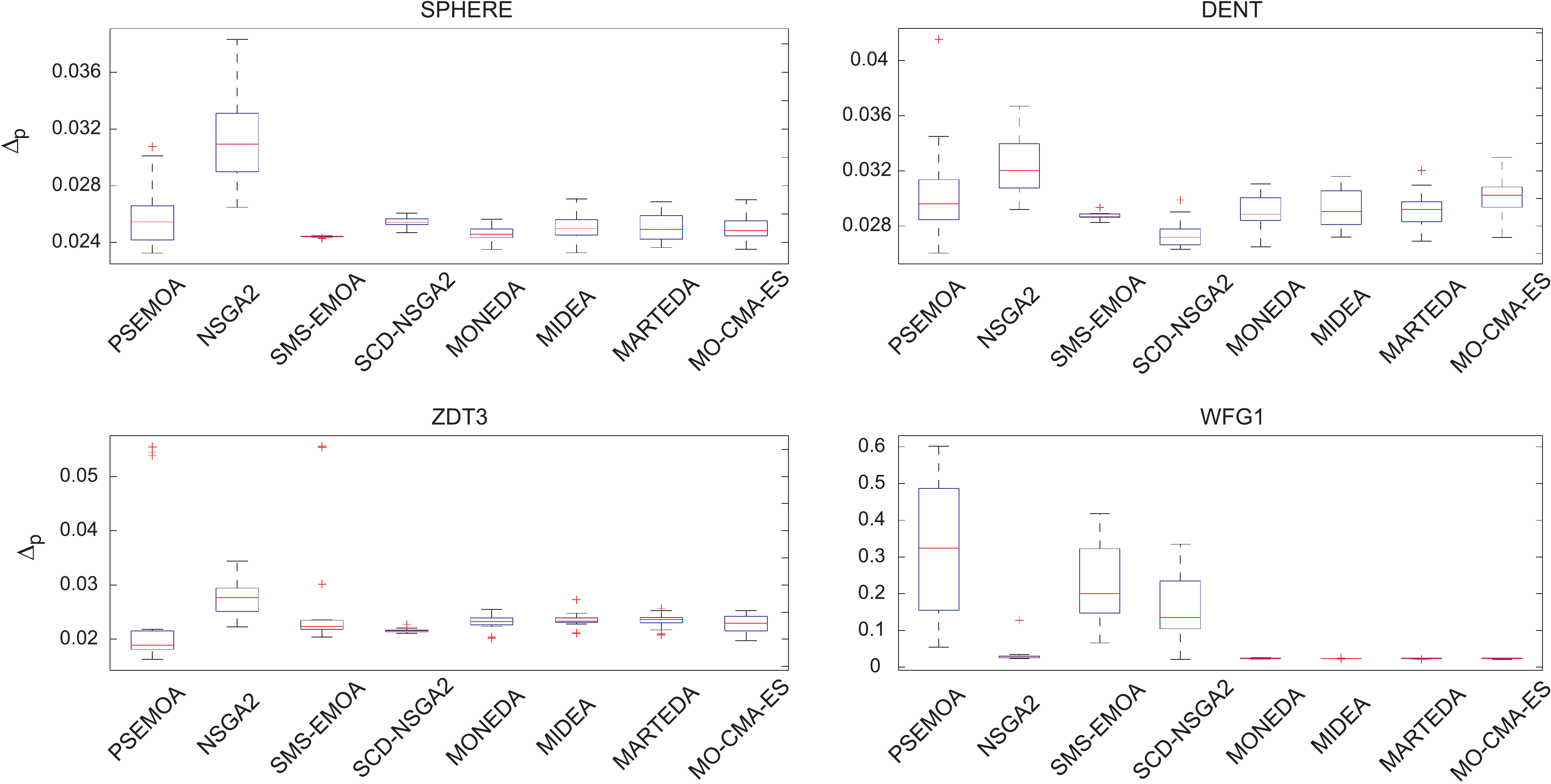}
    \caption{Comparison of MEDAs with EMOAs regarding $\Delta_p$.}
    \label{fig:algos_delta_p}
\end{figure}

At first, overall algorithm behavior is investigated by comparing the MEDAs with the classical EMOA approaches in terms of the Dominated Hypervolume indicator (HV, \cite{ZTL03}) as well as $\Delta_p$ (Figures \ref{fig:algos_hv} and \ref{fig:algos_delta_p}) based on the generated reference fronts as described in Section \ref{subsec:ExSetup}. It becomes obvious that the MEDAs lead to very stable results over the repeated runs which is reflected by both indicators while no MEDA is superior to the others. Moreover, for all test problems the respective performance is comparable to the best performing algorithm out of the classical EMOA set. While over all test problems but WFG1 PSEMOA and SCD-NSGA2 can be considered as best regarding $\Delta_p$, the SMS-EMOA unsurprisingly outperforms the other classical EMOA in terms of HV as it internally optimizes the HV indicator. The test problem WFG1 results in different algorithm rankings within the classical EMOA, i.e. NSGA2 clearly is best regarding both indicators. The stability of MEDA results is very noticeable here as the variability of PSEMOA, SMS-EMOA as well as SCD-NSGA2 is large. Concluding, the MEDAs are at the least competitive with the classical EMOAs on the considered test problems.


\begin{figure*}[htpb]
		\centering
    \includegraphics[width=\textwidth]{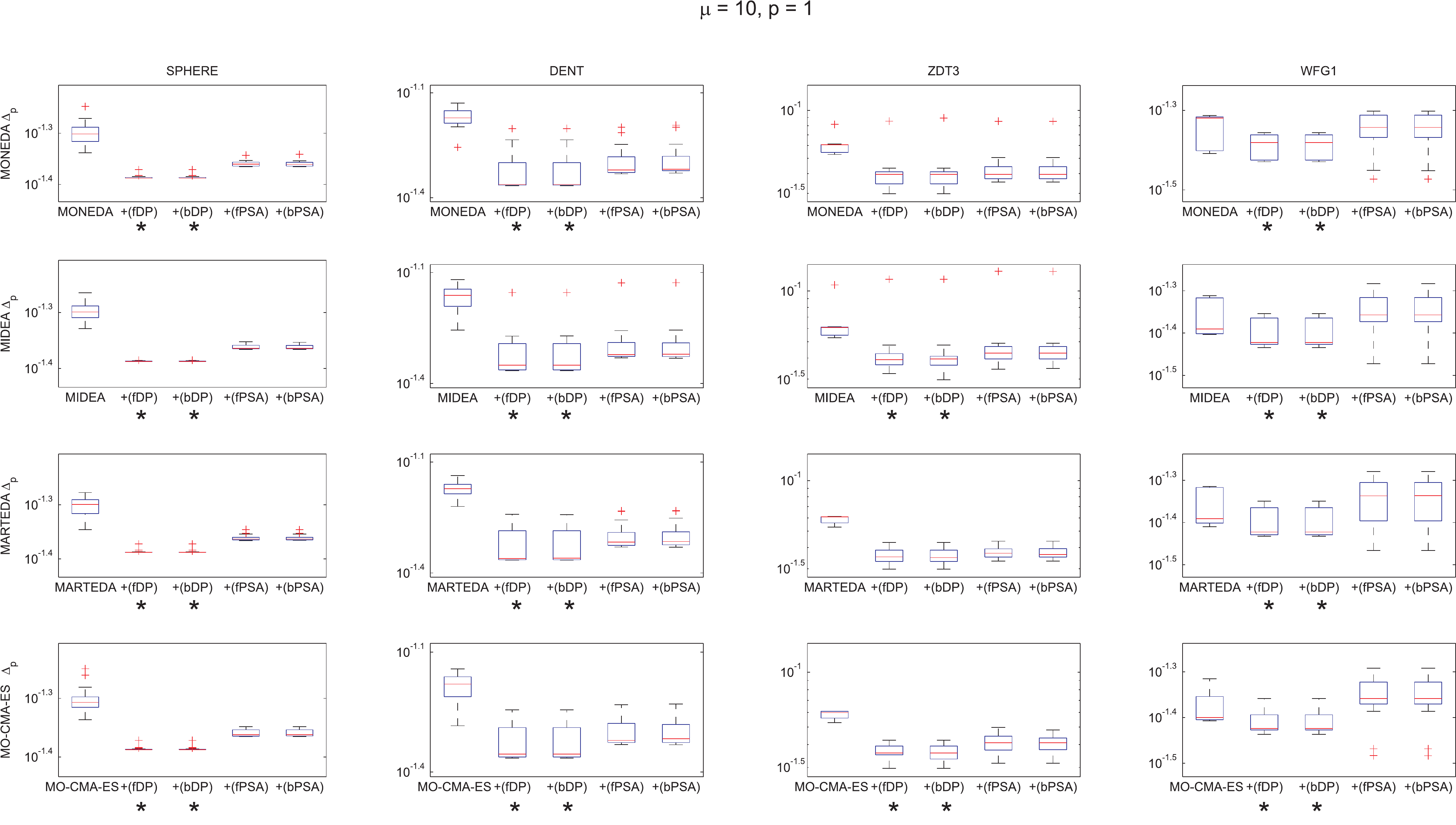}
    \caption{$\Delta_p$- and PSA-postprocessing applied to MEDA results for $\mu = 10$ and $p = 1$.}
    \label{fig:res_p1_10}
\end{figure*}

\begin{figure*}[htpb]
		\centering
    \includegraphics[width=\textwidth]{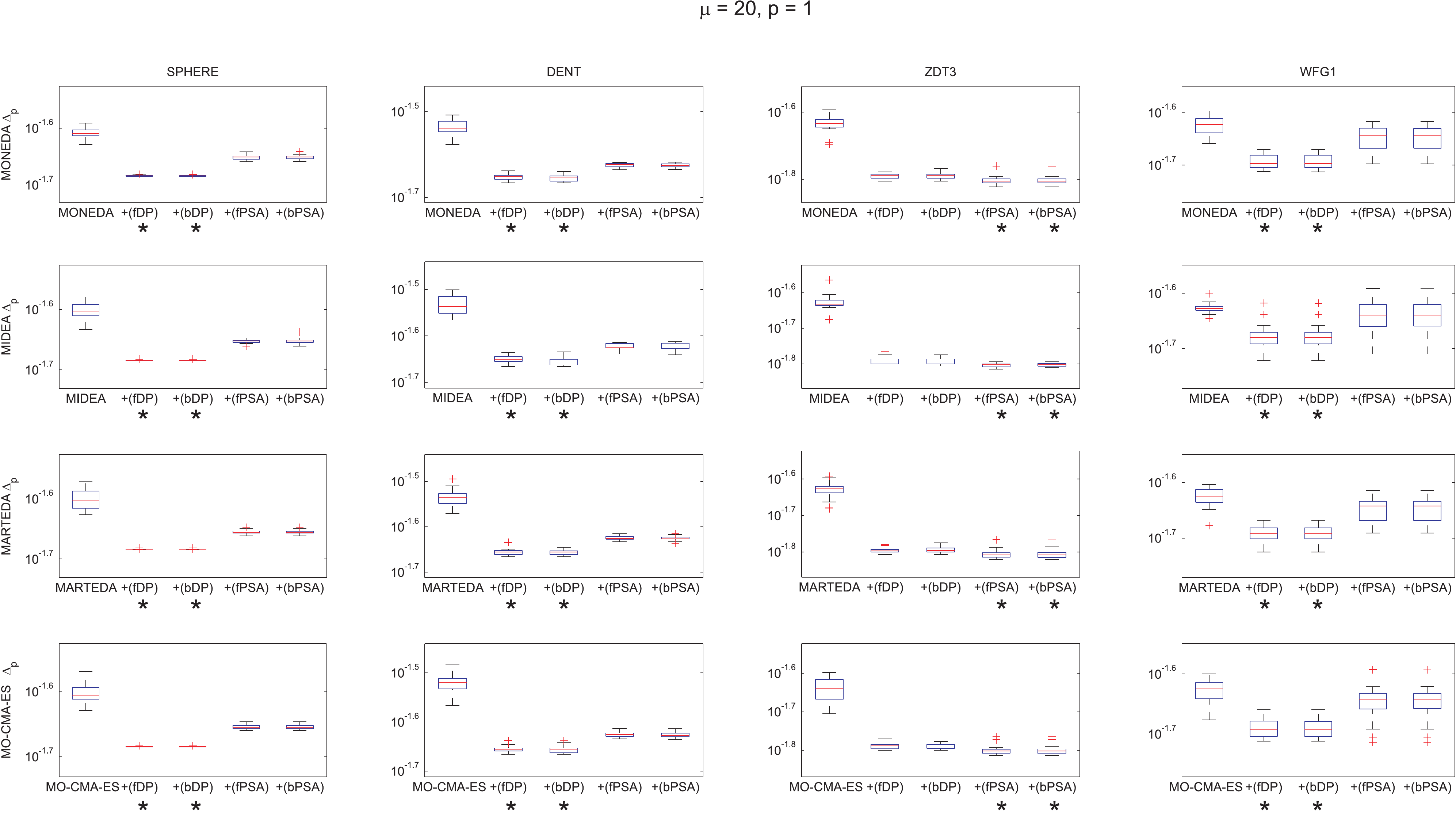}
    \caption{$\Delta_p$- and PSA-postprocessing applied to MEDA results for $\mu = 20$ and $p = 1$.}
    \label{fig:res_p1_20}
\end{figure*}

\begin{figure*}[htpb]
		\centering
    \includegraphics[width=\textwidth]{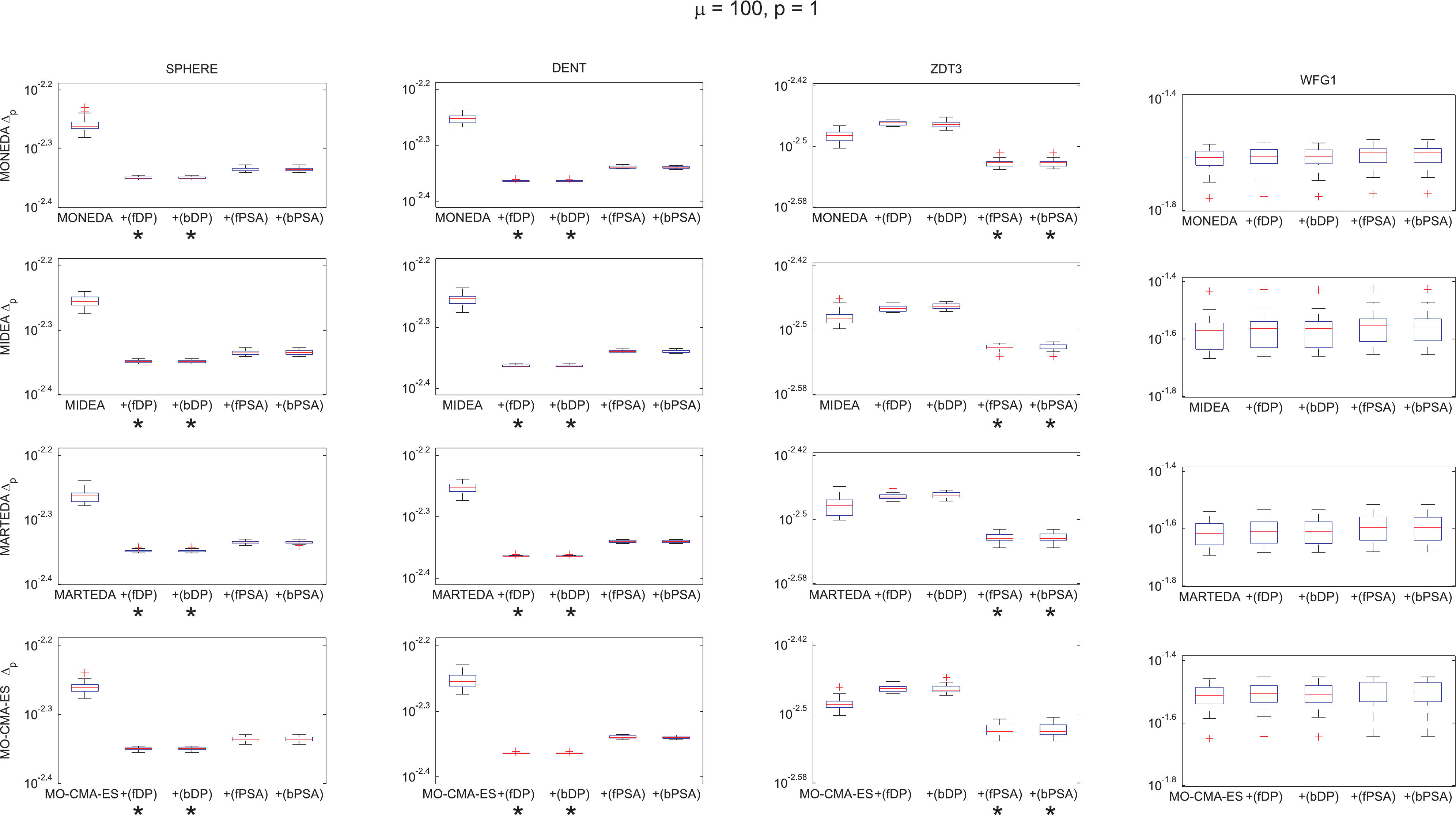}
    \caption{$\Delta_p$- and PSA-postprocessing applied to MEDA results for $\mu = 100$ and $p = 1$.}
    \label{fig:res_p1_100}
\end{figure*}

In Figures \ref{fig:res_p1_10} - \ref{fig:res_p1_100} results of the postprocessing strategies applied to the MEDA variants are visualized in terms of $\Delta_p$ regarding the generated reference fronts for population sizes $\mu \in \{10,20,100\}$. In general, experimental results become more distinguishable with increasing population size but the basic findings are reflected for the smallest population size as well. However, varying $p$ does not have a noticeable effect in this setting, thus the analysis concentrates on $p=1$ for illustration purposes.

Clearly, the postprocessing improves the final approximation quality of the MEDA's run as the primary algorithm in terms of $\Delta_p$ for all population sizes. As the Pareto front of ZDT3 is disconnected, the performance of the considered postprocessing differs from the remaining test problems. Here, the two PSA-based strategies ($+(fPSA)$ and $+(bPSA)$) outperform the archive strategy using linear interpolations of the set of all nondominated points generated in the course of the algorithm run ($+(fDP)$ and $+(bDP)$). This is intuitive as the latter approach ignores the discontinuity in the interpolation and tends to place reference points also in unattainable regions \cite{RTS12}. Thus, this effect increases with the population size as more points have to placed on the linear interpolation of the front. For $\mu=10$ the strategies $+(fDP)$ and $+(bDP)$ are comparable or even slightly better than the PSA-based ones. For all other test problems $+(fDP)$ and $+(bDP)$ generate the largest improvement in terms of $\Delta_p$. Differences are statistically significant in most settings  reflected by a Wilcoxon rank sum test ($p\le 0.05$) and annotated with a $\star$ in Figures \ref{fig:res_p1_10} - \ref{fig:res_p1_100}.

Interestingly, for $\mu=100$ and WFG1 differences between the primary MEDA and all postprocessing variants are very small and not statistically significant.
This is due to the special challenges the test problem poses onto the algorithms. Thorough analysis of the final approximation sets of the MEDAs revealed that those runs do not allow for substantial improvements of the final solution sets as the algorithm performance is not good enough and the archive of solutions does not provide the  means for selecting a better subset regarding $\Delta_p$. However, substantially increasing the budget of the MEDAs will lead to results showing the same tendencies as described above for smaller population sizes. 
As expected, the direction of the archive update  strategy has no influence on the  final approximation quality. However, our experiments revealed that the backward strategy which starts from  the final population of the primary MEDA requires much less updating iterations until the selected subset becomes stable than the forward strategy. Therefore, we recommend to favor this approach over the forward update in general such that the postprocessing will be much more computationally efficient.

\FloatBarrier

\section{Conclusions and Outlook}\label{sec:conc}

Within this work, we first compared the quality of the solutions of four evolutionary multiobjective optimization algorithms (NSGA2, PSEMOA, SCD-NSGA2, SMS-EMOA) and four multiobjective estimation of distribution algorithms (MARTEDA,  MO-CMA-ES, MONEDA, Naive MIDEA) on a set of well-known multiobjective test problems. Based on two considered indicators -- HV and $\Delta_p$ --  it could be shown that the performance of the MEDAs is at least competitive to classical EMOAs. Also, the MEDAs itself performed quite similar to each other.

Due to those competitive and thus promising results, we analyzed whether the performance of the latter ones regarding $\Delta_p$ could even be improved by a subsequent application of either forward or backward $\Delta_p$-update strategies which will lead to a more uniformly distributed solutions along the approximated Pareto front. Experiments
confirmed this hypothesis and apparently the influence of using either one of those update directions is negligible. On the other hand, the $\Delta_p$-based techniques outperform the PSA-based ones besides for problems with disconnected Pareto fronts.

In future work, one might want to improve the MEDAs performance either based on the PSA- or $\Delta_p$-update stra\-te\-gy by integrating that approach into the algorithm's mechanics. This hopefully leads to algorithms, which generate comparable or even better solutions within a single run and therefore without employing any additional postprocessing.

\bibliographystyle{unsrt}
\bibliography{EDA_AHD}

\begin{thebibliography}{10}

\bibitem{dag-2008:mo-inter-evo}
J\"urgen Branke, Kaisa Miettinen, Kalyanmoy Deb, and Roman S{\l}owi\`{n}ski,
  editors.
\newblock {\em Multiobjective Optimization: {I}nteractive and Evolutionary
  Approaches}, volume 5252 of {\em Lecture Notes in Computer Science}.
\newblock Springer Verlag, Berlin/Heidelberg, 2008.

\bibitem{coello-2007:eas-solving-mops}
Carlos~A. Coello~Coello, Gary~B. Lamont, and David~A. Van~Veldhuizen.
\newblock {\em Evolutionary Algorithms for Solving Multi-Objective Problems}.
\newblock Genetic and Evolutionary Computation. Springer, New York, second
  edition, 2007.

\bibitem{Zitzler02}
Eckart Zitzler, Marco Laumanns, Lothar Thiele, Carlos~M. Fonseca, and Viviane
  Grunert~da Fonseca.
\newblock Why quality assessment of multiobjective optimizers is difficult.
\newblock In W.~B. Langdon, E.~Cant\'{u}-Paz, K.~Mathias, R.~Roy, D.~Davis,
  R.~Poli, K.~Balakrishnan, V.~Honavar, G.~Rudolph, J.~Wegener, L.~Bull, M.A.
  Potter, A.C. Schultz, J.F. Miller, E.~Burke, and N.~Jonoska, editors, {\em
  Proceedings of the Genetic and Evolutionary Computation Conference
  (GECCO'2002)}, pages 666--673, San Francisco, California, July 2002. Morgan
  Kaufmann Publishers.

\bibitem{ZTL03}
Eckart Zitzler, Lothar Thiele, Marco Laumanns, Carlos~M. Fonseca, and
  Viviane~Grunert da~Fonseca.
\newblock {Performance assessment of multiobjective optimizers: An analysis and
  review.}
\newblock {\em IEEE Transactions on Evolutionary Computation}, 7(2):117--132,
  2003.

\bibitem{Deb02}
Kalyanmoy Deb, Amrit Pratap, Sameer Agarwal, and T.~Meyarivan.
\newblock A {F}ast and {E}litist {M}ultiobjective {G}enetic {A}lgorithm:
  {NSGA--II}.
\newblock {\em IEEE Transactions on Evolutionary Computation}, 6(2):182--197,
  April 2002.

\bibitem{emmerich:13}
Michael Emmerich, André Deutz, Johannes Kruisselbrink, and PradyumnKumar
  Shukla.
\newblock Cone-based hypervolume indicators: Construction, properties, and
  efficient computation.
\newblock In RobinC. Purshouse, PeterJ. Fleming, CarlosM. Fonseca, Salvatore
  Greco, and Jane Shaw, editors, {\em Evolutionary Multi-Criterion
  Optimization}, volume 7811 of {\em Lecture Notes in Computer Science}, pages
  111--127. Springer Berlin Heidelberg, 2013.

\bibitem{SELC12}
O.~Sch\"utze, X.~Esquivel, A.~Lara, and C.~A. Coello~Coello.
\newblock Using the averaged {H}ausdorff distance as a performance measure in
  evolutionary multiobjective optimization.
\newblock {\em IEEE Transactions on Evolutionary Computation}, 16(4):504--522,
  2012.

\bibitem{GRST11}
K.~Gerstl, G.~Rudolph, O.~Sch{\"u}tze, and H.~Trautmann.
\newblock Finding evenly spaced fronts for multiobjective control via averaging
  {H}ausdorff-measure.
\newblock In {\em Proceedings of 8th International Conference on Electrical
  Engineering, Computing Science and Automatic Control (CCE)}, pages 1--6. IEEE
  Press, 2011.

\bibitem{RTSS13}
G.~Rudolph, H.~Trautmann, S.~Sengupta, and O.~Sch{\"u}tze.
\newblock Evenly spaced {P}areto front approximations for tricriteria problems
  based on triangulation.
\newblock In {\em Proceedings of 7th International Conference on Evolutionary
  Multi-Criterion Optimization (EMO 2013)}, pages 443--458, Berlin Heidelberg,
  2013. Springer, LNCS 7811.

\bibitem{DRST13}
C.~Dominguez-Medina, G.~Rudolph, O.~Sch{\"u}tze, and H.~Trautmann.
\newblock Evenly spaced {P}areto fronts of quad-objective problems using {PSA}
  partitioning technique.
\newblock In {\em Proceedings of 2013 IEEE Congress on Evolutionary Computation
  (CEC 2013)}, pages 3190--3197, Piscataway (NJ), 2013. IEEE Press.

\bibitem{corne-2008:talk}
David~Wolfe Corne.
\newblock Single objective = past, multiobjective = present, ??? = future.
\newblock In Zbigniew Michalewicz, editor, {\em 2008 IEEE Conference on
  Evolutionary Computation (CEC)}, Piscataway, New Jersey, 2008. IEEE Press.

\bibitem{michalski-2000:lem}
R.~S. Michalski.
\newblock Learnable evolution model: {E}volutionary processes guided by machine
  learning.
\newblock {\em Machine Learning}, 38:9--40, 2000.

\bibitem{garrett-2008:mo-fit-land}
J.~Deon Garrett.
\newblock {\em Multiobjective Fitness Landscape Analysis and the Design of
  Effective Memetic Algorithms}.
\newblock PhD thesis, Memphis State University, Memphis, TN, USA, 2008.
\newblock AAI3310115.

\bibitem{verel-2010:mofitland}
S{\'e}bastien Verel, Laetitia Jourdan, Clarisse Dhaenens, and Arnaud Liefooghe.
\newblock Set-based multiobjective fitness landscapes: definition, properties.
\newblock In {\em 4th Workshop on Theory of Randomized Search Heuristics},
  Paris, France, March 2010.

\bibitem{humeau-2013:paradiseo-mo}
J{\'e}r{\'e}mie Humeau, Arnaud Liefooghe, El-Ghazali Talbi, and S{\'e}bastien
  Verel.
\newblock Paradiseo-{MO}: {F}rom fitness landscape analysis to efficient local
  search algorithms.
\newblock {\em Journal of Heuristics}, 19(6):881--915, 2013.

\bibitem{lozano-2005:edas}
J.~A. Lozano, P.~Larra{\~{n}}aga, I.~Inza, and E.~Bengoetxea, editors.
\newblock {\em Towards a New Evolutionary Computation: {A}dvances on Estimation
  of Distribution Algorithms}.
\newblock Springer Verlag, 2006.

\bibitem{pelikan-2006:medas}
Martin Pelikan, Kumara Sastry, and David~E. Goldberg.
\newblock Multiobjective estimation of distribution algorithms.
\newblock In Martin Pelikan, Kumara Sastry, and Erick Cant\'{u}-Paz, editors,
  {\em Scalable Optimization via Probabilistic Modeling: From Algorithms to
  Applications}, Studies in Computational Intelligence, pages 223--248.
  Springer Verlag, 2006.

\bibitem{bosman2005MIDEA}
Peter~AN Bosman and Dirk Thierens.
\newblock The naive midea: A baseline multi-objective ea.
\newblock In {\em Evolutionary Multi-Criterion Optimization}, pages 428--442.
  Springer, 2005.

\bibitem{hansen-2003:cma-es}
N.~Hansen, S.D. Muller, and P.~Koumoutsakos.
\newblock Reducing the time complexity of the derandomized evolution strategy
  with covariance matrix adaptation ({CMA--ES}).
\newblock {\em Evolutionary Computation}, 11(1):1--18, 2003.

\bibitem{beyer-2002:evol-strats}
Hans-Georg Beyer and Hans-Paul Schwefel.
\newblock Evolution strategies --- {A} comprehensive introduction.
\newblock {\em Natural Computing: an international journal}, 1(1):3--52, 2002.

\bibitem{igel2007MOCMAES}
Christian Igel, Nikolaus Hansen, and Stefan Roth.
\newblock Covariance matrix adaptation for multi-objective optimization.
\newblock {\em Evolutionary computation}, 15(1):1--28, 2007.

\bibitem{ahn-2007:diversity-preservation}
Chang~Wook Ahn and R.~S. Ramakrishna.
\newblock Multiobjective real-coded {B}ayesian optimization algorithm
  revisited: {D}iversity preservation.
\newblock In {\em GECCO '07: Proceedings of the 9th annual conference on
  Genetic and evolutionary computation}, pages 593--600, New York, NY, USA,
  2007. ACM Press.

\bibitem{yuan-2005:diversity-eda}
Bo~Yuan and Marcus Gallagher.
\newblock On the importance of diversity maintenance in estimation of
  distribution algorithms.
\newblock In {\em {GECCO} '05: {P}roceedings of the 2005 conference on Genetic
  and evolutionary computation}, pages 719--726, New York, NY, USA, 2005. ACM
  Press.

\bibitem{shapiro-2006:diversity-eda}
Jonathan Shapiro.
\newblock Diversity loss in general estimation of distribution algorithms.
\newblock In {\em Parallel Problem Solving from Nature - PPSN IX}, pages
  92--101, 2006.

\bibitem{marti-2011:phd-thesis}
Luis Mart\'{i}.
\newblock {\em Scalable Multi-Objective Optimization}.
\newblock PhD thesis, Departmento de Inform{\'a}tica, Universidad Carlos III de
  Madrid, Colmenarejo, Spain, 2011.

\bibitem{Bosman2007}
Peter~A.N. Bosman and Dirk Thierens.
\newblock {Adaptive variance scaling in continuous multi-objective
  estimation-of-distribution algorithms}.
\newblock In {\em Proceedings of the 9th annual conference on Genetic and
  evolutionary computation - GECCO '07}, page 500, New York, New York, USA,
  2007. ACM Press.

\bibitem{Bosman2010}
Peter A.~N. Bosman.
\newblock The anticipated mean shift and cluster registration in mixture-based
  {EDAs} for multi-objective optimization.
\newblock In {\em Proceedings of the 12th annual conference on Genetic and
  evolutionary computation - GECCO '10}, page 351, New York, New York, USA,
  2010. ACM Press.

\bibitem{marti2008MONEDA}
Luis Mart{\'\i}, Jes{\'u}s Garc{\'\i}a, Antonio Berlanga, and Jos{\'e}~Manuel
  Molina.
\newblock Introducing moneda: Scalable multiobjective optimization with a
  neural estimation of distribution algorithm.
\newblock In {\em Proceedings of the 10th Annual Conference on Genetic and
  Evolutionary Computation}, pages 689--696. ACM, 2008.

\bibitem{marti-2011:mb-gng-orl}
Luis Mart\'{i}, Jes\'{u}s Garc\'{i}a, Antonio Berlanga, Carlos~A.
  Coello~Coello, and Jos\'{e}~Manuel Molina.
\newblock {MB-GNG}: {A}ddressing drawbacks in multi-objective optimization
  estimation of distribution algorithms.
\newblock {\em Operations Research Letters}, 39(2):150--154, 2011.

\bibitem{grossberg-1982:match-based}
Stephen Grossberg.
\newblock {\em Studies of Mind and Brain: {N}eural Principles of Learning,
  Perception, Development, Cognition, and Motor Control}.
\newblock Reidel, Boston, 1982.

\bibitem{marti2011MARTEDA}
Luis Mart{\'\i}, Jes{\'u}s Garc{\'\i}a, Antonio Berlanga, and Jos{\'e}~M
  Molina.
\newblock Multi-objective optimization with an adaptive resonance theory-based
  estimation of distribution algorithm: a comparative study.
\newblock In {\em Learning and Intelligent Optimization}, pages 458--472.
  Springer, 2011.

\bibitem{brake-2007:sampling-edas}
J{\"u}ergen Branke, Clemens Lode, and Jonathan~L. Shapiro.
\newblock Addressing sampling errors and diversity loss in umda.
\newblock In {\em Proceedings of the 9th annual conference on Genetic and
  evolutionary computation --- GECCO '07}, pages 508--515. ACM Press, 2007.

\bibitem{Muhlenbein_and_Paas:1996}
H.~M{\"{u}}hlenbein and G.~Paa{\ss}.
\newblock From recombination of genes to the estimation of distributions {I}.
  {B}inary parameters.
\newblock In Hans-Michael Voigt, Werner Ebeling, Ingo Rechenberg, and Hans-Paul
  Schwefel, editors, {\em Parallel {P}roblem {S}olving from {N}ature - {PPSN
  IV}}, pages 178--187, Berlin, 1996. Springer Verlag.
\newblock {LNCS} 1141.

\bibitem{pena-2004:ga-eda}
J.~M. Pena, Victor Robles, Pedro Larra\~{n}aga, V.~Herves, F.~Rosales, and
  M.~S. P\'{e}rez.
\newblock {GA-EDA}: {H}ybrid evolutionary algorithm using genetic and
  estimation of distribution algorithms.
\newblock In {\em Innovations in Applied Artificial Intelligence}, pages
  361--371, Heidelberg/Berlin, 2004. Springer.

\bibitem{zhang-2005:div-loss}
Q.~Zhang, Jianyong Sun, and Edward Tsang.
\newblock An evolutionary algorithm with guided mutation for the maximum clique
  problem.
\newblock {\em IEEE Transactions on Evolutionary Computation}, 9(2):192--200,
  April 2005.

\bibitem{durillo2011jmetal}
Juan~J Durillo and Antonio~J Nebro.
\newblock jmetal: A java framework for multi-objective optimization.
\newblock {\em Advances in Engineering Software}, 42(10):760--771, 2011.

\bibitem{huband2006review}
Simon Huband, Philip Hingston, Luigi Barone, and Lyndon While.
\newblock A review of multiobjective test problems and a scalable test problem
  toolkit.
\newblock {\em Evolutionary Computation, IEEE Transactions on}, 10(5):477--506,
  2006.

\bibitem{beume2007sms}
Nicola Beume, Boris Naujoks, and Michael Emmerich.
\newblock Sms-emoa: Multiobjective selection based on dominated hypervolume.
\newblock {\em European Journal of Operational Research}, 181(3):1653—1669,
  2007.

\bibitem{kukkonen2006improved}
Saku Kukkonen and Kalyanmoy Deb.
\newblock Improved pruning of non-dominated solutions based on crowding
  distance for bi-objective optimization problems.
\newblock In {\em Proceedings of the World Congress on Computational
  Intelligence (WCCI-2006)(IEEE Press). Vancouver, Canada}, pages 1179--1186,
  2006.

\bibitem{RTS12}
G.~Rudolph, H.~Trautmann, and O.~Sch\"utze.
\newblock {Homogene Approximation der Paretofront bei mehrkriteriellen
  Kontrollproblemen}.
\newblock {\em Automatisierungstechnik (at)}, 60(10):612--621, 2012.

\end{thebibliography}

\end{document}